\documentclass[letterpaper, 10 pt, conference]{ieeeconf}  % Comment this line out if you need a4paper
\IEEEoverridecommandlockouts                              % This command is only needed if
\usepackage[colorlinks,linkcolor=blue]{hyperref}      % hyperlinks
\usepackage{url}            % simple URL typesetting
\usepackage{booktabs}       % professional-quality tables
\usepackage{amsfonts}       % blackboard math symbols
\usepackage{nicefrac}       % compact symbols for 1/2, etc.
\usepackage{microtype}      % microtypography
\usepackage[noadjust]{cite}
\usepackage{amsmath,amssymb}
\usepackage{gensymb}
\usepackage[ruled,linesnumbered]{algorithm2e}
\usepackage{color}
\usepackage{amsfonts}
\usepackage{subfig}
\usepackage{graphicx}
\usepackage{tabulary,multirow,overpic,xcolor}
\usepackage{wrapfig}
\usepackage{sidecap}
\usepackage{todonotes}
\usepackage{blindtext}
\usepackage{amsfonts,amssymb}
\usepackage{makecell}
\usepackage{epstopdf}
\usepackage{multirow}

\hypersetup{
    colorlinks=true,
    linkcolor=blue,
    filecolor=magenta,
    urlcolor=blue,
}

\newcommand{\app}{\raise.17ex\hbox{$\scriptstyle\sim$}}
\newcolumntype{x}[1]{>{\centering\arraybackslash}p{#1pt}}

\newlength\savewidth

\makeatletter\renewcommand\paragraph{\@startsection{paragraph}{4}{\z@}
  {.5em \@plus1ex \@minus.2ex}{-.5em}{\normalfont\normalsize\bfseries}}\makeatother

\title{\LARGE \bf Vision-Based Defect Classification and Weight Estimation of Rice Kernels}

\author{Xiang Wang$^{1}$, {Kai Wang}$^{1}$, {Xiaohong Li}$^{1}$, and {Shiguo Lian}$^{1}$% \\
\thanks{ $^{1}$All authors are with AI Innovation and Application Center, China Unicom.
{\tt\small wangx386, wangk115, lixh585, liansg@chinaunicom.cn}}}
\begin{document}
%\history{Date of publication xxxx 00, 0000, date of current version xxxx 00, 0000.}
%\doi{10.1109/ACCESS.2017.DOI}

\maketitle
\title{Vision-Based Defect Classification and Weight Estimation of Rice Kernels}

%\author{\uppercase{Xiang Wang}\authorrefmark{1},
%\uppercase{Kai Wang\authorrefmark{1}, Xiaohong Li\authorrefmark{1}, and Shiguo Lian}\authorrefmark{1}, \IEEEmembership{Member, IEEE}}
%\thanks{AI Innovation and Application Center, China Unicom, Beijing, China}

%\markboth
%{Author \headeretal: Preparation of Papers for IEEE TRANSACTIONS and JOURNALS}
%{Author \headeretal: Preparation of Papers for IEEE TRANSACTIONS and JOURNALS}

%\corresp{Corresponding author: Shiguo Lian (e-mail: liansg@chinaunicom.cn).}
.

%\titlepgskip=-15pt

\begin{abstract}
Rice is one of the main staple food in many areas of the world. The quality estimation of rice kernels are crucial in terms of both food safety and socio-economic impact. This was usually carried out by quality inspectors in the past, which may result in both objective and subjective inaccuracies. In this paper, we present an automatic visual quality estimation system of rice kernels, to classify the sampled rice kernels according to their types of flaws, and evaluate their quality via the weight ratios of the perspective kernel types. To compensate for the imbalance of different kernel numbers and classify kernels with multiple flaws accurately, we propose a multi-stage workflow which is able to locate the kernels in the captured image and classify their properties. We define a novel metric to measure the relative weight of each kernel in the image from its area, such that the relative weight of each type of kernels with regard to the all samples can be computed and used as the basis for rice quality estimation. Various experiments are carried out to show that our system is able to output precise results in a contactless way and replace tedious and error-prone manual works.
\end{abstract}


\section{Introduction}
\label{sec:Intro}

Rice is the staple food for people in many parts of the world, and its quality is related to food health as well as the economic interests of agricultural dealers, who make offer to the farmers according to the quality of the collected rice kernels. Manual rice quality estimation is quite tedious, as it requires experienced inspectors to identify and pick up the kernels with various defects one by one and weigh them carefully. The precision of the result is subject to the skill and conscientiousness of the inspectors.

With the development of computer vision technique in recent decades, lots of attempts have been made to classify rice types and defects automatically~\cite{AAK21}. A wide range of computer vision algorithms, from traditional geometric~\cite{ajay2013quality} to deep-learning based methods~\cite{kiratiratanapruk2020using} have been utilized. However, some inherent problems in precise rice quality analysis still exist. For example, rice kernels have various properties divided according to different features, and there are even some kernels containing more than one kinds of defects. In the past, the classification work often relies on the back and forth manual checks of the inspectors. Meanwhile, the estimation of weight ratios of each type of kernel, which is used as the key metric for rice quality and was usually measured via high-precision digital steelyards, is not easy to perform only through its appearance. The reason behind is that although state-of-the-art artificial intelligence has been able to achieve remarkable results in various fields, manual rules and post-processing works are still indispensable in practice to make it applicable enough to replace skilled engineers and workers.

\begin{figure}[htbp]
\centerline{\includegraphics[width=0.46\textwidth]{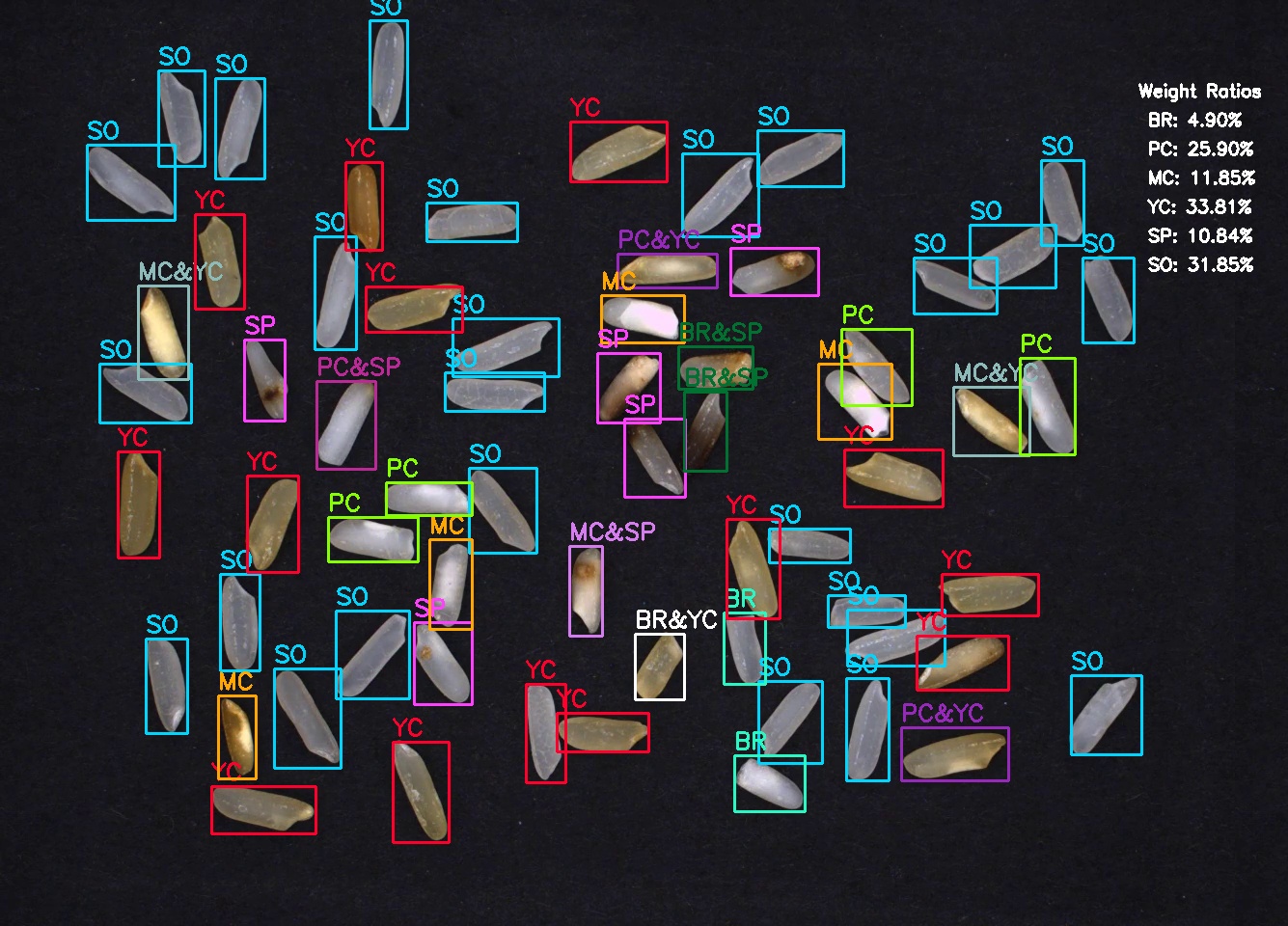}}
\caption{Rice quality estimation results with the proposed system. The type of each rice kernel is labeled on its bounding box, and the weight ratios are shown in the upper right area. PC, MC, YC, SP, BR and SO represent Partial-Chalky, Mass-Chalky, Yellow-Colored, Spotted, Broken and Sound kernels respectively.}
\label{fig:res_all}
\end{figure}

To solve these problems which existing vision-based methods are not able to handle, we propose a novel approach for automatic quality estimation of rice kernels. We designed a hardware system to obtain high-quality images of rice kernels with little lighting interference and background noise. To detect and classify the kernels with various types of defects, we employed a multi-stage classification approach to perform multi-classification of rice kernel flaws, such that a single kernel with dual defects can be detected and the classification accuracy for different defects is improved. Then we use a vision-based method to estimate the relative weights of kernels with each type of defect, by measuring its density via a novel weight-per-pixel metric and multiplying it with the segmented areas. An example of our result is shown in Fig.~\ref{fig:res_all}. With such workflows, the quality of the rice kernels can be automatically analyzed and presented to the inspectors, and tedious manual works can be replaced.

% contribution summary
The contributions of this paper include:

\begin{itemize}
    \item The introduction of a multi-stage method for detection and classification of rice kernels with various defects;
    \item The proposal of a weight-per-pixel metric for the estimation of weight ratios of each type of kernels;
    \item The design of a complete system which integrates these approaches to produce high-precision rice quality evaluation results automatically.
\end{itemize}

% paper constructure
The rest of the paper is organized as follows: Section~\ref{sec:Review} reviews existing methods for rice classification and defect detection. Section~\ref{sec:method} introduces the details of the proposed system and methods. Experimental evaluation is presented in Section~\ref{sec:Results}, and Section~\ref{sec:Con} gives the conclusions.

\section{Related Work}
\label{sec:Review}
During the last decades, various approaches for automatic rice classification and quality analysis have been proposed.

\textbf{Rice classification:} Most work adopts the following route: extracting features with image processing algorithms and then classifying the rice based on these features. Kuo \emph{et al.}~\cite{kuo2016identifying} classified 30 varieties rice grains using image processing and sparse-representation-based classification (SRC). Kambo and Yerpude~\cite{kambo2014classification} distinguished the variety of Basmati Rice Grain using K-NN and Principal Component Analysis (PCA), after preprocessing the images with smoothing and segmentation techniques. To improve the classification accuracy, Rad \emph{et al.}~\cite{mousavirad2012design} presented a rice classification algorithm with optimal morphological features and back propagation neural network-based (BPNN). Here, 18 morphological features were extracted, and 6 features were selected. Silva and Sonnadara~\cite{silva2013classification} combined neural network (NN) with PCA to classify the rice seed varieties. 34 features were extracted by some pre-processing operations before PCA was applied to perform dimensionality reduction, and one individual neural network was created for each feature set.

With the widespread application of deep learning and their excellent performance in image tasks, more and more rice classification work starts to adopt deep neural networks. Lin \emph{et al.}~\cite{lin2017determination} proposed a model using convolutional neural networks (CNN) for rice kernel classification which reached a 99.52$\%$ accuracy. However, they found the accuracy of classification is closely related to the preprocessing effect of the image, which was done with image enhancement operations, such as re-scaling, mean subtraction, and feature standardization. Qiu \emph{et al.}~\cite{qiu2018variety} presented a hyperspectral-CNN based rice variety classification algorithm. The rice image captured from the hyperspectral system was pre-processed with a wavelet transform and image segmentation process. 100-3000 rice samples were used to build KNN, SVM, and CNN models. They found CNN outperformed SVM and KNN. Similar to this work, Chatnuntawech \emph{et al.}~\cite{chatnuntawech2018rice} provided a rice classification algrithom for identification with the synergy between hyperspectral imaging and deep CNN. They found that the classification effect has been significantly improved with the deepening of the number of neural network layers. Patel proposed two methods for rice types classification. One used CNN with segmented rice images as input. Another used a pretrained VGG-16 model and transfer learning to achieve a better result. Kiratiratanapruk \emph{et al.}~\cite{kiratiratanapruk2020using} applied deep learning to detect and identify rice disease in images. They conducted experiments with 4 models namely Faster R-CNN~\cite{ren2015faster}, RetinaNet~\cite{lin2017focal}, YOLOv3~\cite{yolov5} and Mask RCNN~\cite{he2017mask}. %It was reported that YOLOv3 provided the best performance.

\textbf{Rice quality analysis:}
Different solutions have been applied on rice grain analysis. These approaches can broadly be classified into geometric, statistical, and machine learning.

Geometric approaches consider morphological features as key factors to analysis. Ajay \emph{et al.}~\cite{ajay2013quality} used shape descriptors and geometric features to determine quantity of broken kernels among milled rice samples. Asif \emph{et al.}~\cite{asif2018rice} used morphological features to determine the quality of five types of rice grains after a grain classification. Mahale and Korde~\cite{mahale2014rice} applied image processing techniques to grade and evaluate rice grains based on grain size and shape, such as length, width and their ratio. In contrast, Ali \emph{et al.}~\cite{ali2017low} proposed an low cost solution for rice quality analysis based on more features. They computed the average length, average width, the area and number of small rice grains, medium rice grains and broken rice grains to analysis rice quality.

Statistical approaches primarily focus on summarizing data and making inferences according to the population. Mahajan and Kaur~\cite{mahajan2014quality} proposed a method of quality analysis for three types of Indian Basmati rice grains including normal grains, long grains and small grains. They applied morphological closing and opening operations and top-hat transformation to calculate the length of the major and minor axes of rice. The rice was graded by analyzing histograms.

With the successful application of machine learning and its excellent performance, Agustin and Oh~\cite{agustin2008automatic} proposed an automatic quality evaluation framework for milled rice kernels, in which a probabilistic neural network (PNN) classifier is used to detect defective rice including head rice, broken, and brewer kernels. At the same time, a linear regression model was developed for estimating individual kernel weight with a given blob area.~\cite{verma2010image} applied neural network and image processing approach for rice grain identification and grading on three varieties of Indian rice. Rice images, obtained from a flatbed scanner, were pre-processed with several image smoothing operations. The length, width, and perimeter of the rice grains were extracted and input neural network to do classification. It was able to accurately classify rice into sound, cracked, chalky, broken and damaged kernels. Ngampak and Piamsa-Nga~\cite{ngampak2015image} proposed a method for finely classifying broken rice into small broken, broken, big broken and head rice.  They used Least-Square Support Vector Machine (LS-SVM) with Radius Basis Function (RBF) kernel as their classifier. Kaur \emph{et al.}~\cite{kaur2013classification} divided rice kernels into four grades using Multi-Class SVM. They categorized rice into head rice, broken rice and brewers according to the kernel shape, length and chalkiness.

While there are many different methods for detecting rice with different defects, none of them have classified rice with dual properties. %Although ~\cite{agustin2008automatic} proposed a linear regression model to estimate kernel weight, their method fails in some cases.

\section{Method}
\label{sec:method}
In this section, the rice kernel quality evaluation problem will be formulated first. Then the hardware setup for rice image acquisition and the methods for rice kernel classification and weight ratio estimation will be introduced in detail.

%first give a detailed introduction on the proposed grid-based NMS sample strategy, which is used make a better use of informative features on metric learning, thus lease the discreteness between features and descriptors. Then we descriptor the definition and application of "uncertainty" along with two ways to generate uncertainty from network or training image pair.

\subsection{Problem Statement}
\label{sec:method:ps}

\begin{figure}[htbp]
\captionsetup[subfloat]{labelformat=empty}
	\centering
    \subfloat[PC]{
		\begin{minipage}[t]{0.12\linewidth}
			\centering
			\includegraphics[width=1\linewidth]{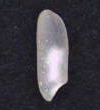}
		\end{minipage}
	}
	\subfloat[MC]{
		\begin{minipage}[t]{0.12\linewidth}
			\centering
			\includegraphics[width=1\linewidth]{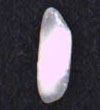}
		\end{minipage}
	}
	\subfloat[YC]{
		\begin{minipage}[t]{0.12\linewidth}
			\centering
			\includegraphics[width=1\linewidth]{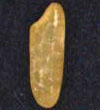}
		\end{minipage}
	}
    \subfloat[SP]{
		\begin{minipage}[t]{0.12\linewidth}
			\centering
			\includegraphics[width=1\linewidth]{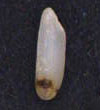}
		\end{minipage}
	}
	\subfloat[BR]{
		\begin{minipage}[t]{0.12\linewidth}
			\centering
			\includegraphics[width=1\linewidth]{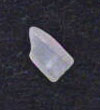}
		\end{minipage}
	}
	\subfloat[SO]{
		\begin{minipage}[t]{0.12\linewidth}
			\centering
			\includegraphics[width=1\linewidth]{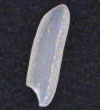}
		\end{minipage}
	}\\
    \subfloat[PC\&YC]{
		\begin{minipage}[t]{0.12\linewidth}
			\centering
			\includegraphics[width=1\linewidth]{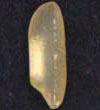}
		\end{minipage}
	}
	\subfloat[MC\&YC]{
		\begin{minipage}[t]{0.12\linewidth}
			\centering
			\includegraphics[width=1\linewidth]{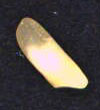}
		\end{minipage}
	}
	\subfloat[BR\&YC]{
		\begin{minipage}[t]{0.12\linewidth}
			\centering
			\includegraphics[width=1\linewidth]{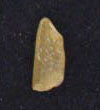}
		\end{minipage}
	}
    \subfloat[PC\&SP]{
		\begin{minipage}[t]{0.12\linewidth}
			\centering
			\includegraphics[width=1\linewidth]{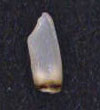}
		\end{minipage}
	}
	\subfloat[MC\&SP]{
		\begin{minipage}[t]{0.12\linewidth}
			\centering
			\includegraphics[width=1\linewidth]{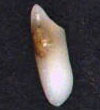}
		\end{minipage}
	}
	\subfloat[BR\&SP]{
		\begin{minipage}[t]{0.12\linewidth}
			\centering
			\includegraphics[width=1\linewidth]{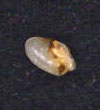}
		\end{minipage}
	}
	\caption{Example images of different rice kernel types. Top row: rice kernels with single property. Bottom row: rice kernels with dual properties.}
	\label{fig:all_classes}
\end{figure}

In this paper, we focus on the detection of the main types of kernel flaws for new rice, including Partial-Chalky kernels (PC), Mass-Chalky kernels (MC), Yellow-Colored kernels (YC), Spotted kernels (SP), and Broken kernels (BR). A chalky kernel refers to a kernel with opaque white parts in endosperm. Depending on the percentage of this area with regard to the whole kernel, it is classified into PC ($\leq 50\%$) and MC ($> 50\%$). As its name suggests, a YC kernel refers to a kernel whose body is yellow. If the rice kernel has disease spot on the surface, it is regarded as a spotted kernel. Broken kernels are small pieces or particles of kernels whose length are less than $2/3$ of the average length of the whole kernels. If a kernel has none of the above defects, it is considered as a perfect kernel, also known as a Sound Kernel (SO).

It is also observed that some rice kernels could have dual properties: a partial-chalky kernel may also be yellow-colored, a spotted kernel may be incomplete (a broken kernel), etc. Common property combinations include: Partial-Chalky and Yellow-Colored (PC\&YC), Mass-Chalky and Yellow-Colored (MC\&YC), Broken and Yellow-Colored (BR\&YC), Partial-Chalky and Spotted (PC\&SP), Mass-Chalky and Spotted (MC\&SP), Broken and Spotted(BR\&SP). Kernels with other flaw combinations are rarely seen and thus can be ignored.

%This is not all possible combinations, just the ones that may be faced by our customers. Because some combinations basically do not appear in reality, for example, a new grain of rice is both yellow and diseased. Furthermore, the customer defines some rules, such as it is not concerned with the presence of chalky parts for broken kernels.

We need to classify the rice kernels with one of the above-mentioned single or dual properties, and the rice quality is estimated by computing the ratio of the weight of each type of kernels with regard to the sum of them. Note that only the weight ratio of kernels with each single types will be calculated, and the weight of a kernel with dual properties will be summed up to the total weights of kernels with each of the two properties respectively.

The weight ratio can thus be calculated as follows. Given a set of rice kernels $\{K_i\}$, where $i\in$ $\{$PC, MC, YC, SP, BR, SO, PC\&YC, MC\&YC, BR\&YC, PC\&SP, MC\&SP, BR\&SP$\}$, the total weight of all kernels is $\sum{W_i}$ (where $W_i$ refers to the weight of the kernel type $i$), and the weight ratio $R_t$ of each type of rice kernel, where $t\in$ $\{$PC, MC, YC, SP, BR, SO$\}$, is formulated as:

\begin{equation}\label{eq:ratio}
\begin{aligned}
&R_{PC}= \frac{W_{PC}+W_{PC\&YC}+W_{PC\&SP}}{\sum W_i},\\
&R_{MC}= \frac{W_{MC}+W_{MC\&YC}+W_{MC\&SP}}{\sum W_i},\\
&R_{YC}= \frac{W_{YC}+W_{PC\&YC}+W_{MC\&YC}+W_{BR\&YC}}{\sum W_i},\\
&R_{SP}= \frac{W_{SP}+W_{PC\&SP}+W_{MC\&SP}+W_{SP\&BR}}{\sum W_i},\\
&R_{BR}= \frac{W_{BR}+W_{BR\&YC}+W_{BR\&SP}}{\sum W_i},\\
&R_{SO}= \frac{W_{SO}}{\sum W_i}.
\end{aligned}
\end{equation}

As the weights of kernels with dual properties are calculated twice, the sum of all $R_t$ will probably not be 1.

\subsection{Hardware Setup}
\label{sec:method:hard}

\begin{figure}[htbp]
    \centering
    \begin{tabular}{c@{\hskip3pt}c@{\hskip3pt}c}
         \includegraphics[width=0.48\textwidth]{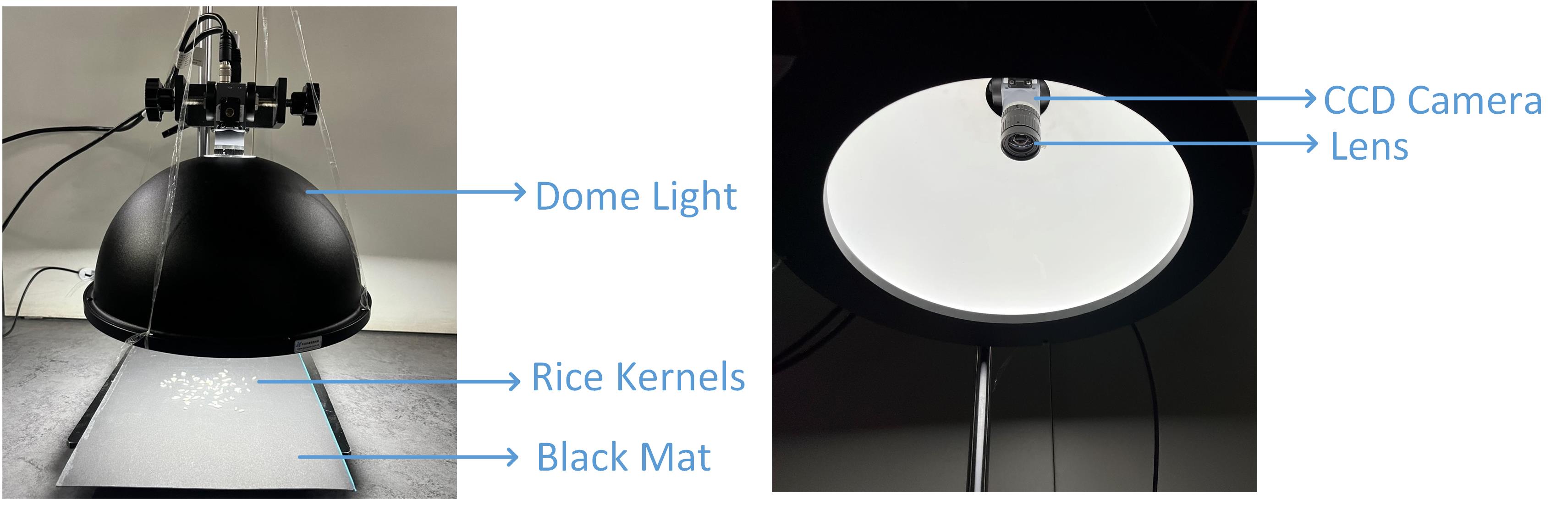}
    \end{tabular}
    \caption{The outside (left) and inside (right) views of the hardware system.}
    \label{fig:hard}
\end{figure}

The hardware system for capturing the images of rice kernels is shown in Fig.~\ref{fig:hard}. We used a white dome light as the light source to make the illumination evenly distributed and meanwhile avoid the influence of outside lightings. A black mat with rough material is used as the background to avoid light reflections. The kernels are placed on the mat manually by scattering through a griddle, such that they can be evenly distributed and have few overlaps. A megapixel CCD camera and a lens with short focus length are used to capture the images of the kernels. The resolution of each image is $1280*960$. All these devices are fixed on a scaffold and are thus stable enough during the measurement process.

The light source and camera are connected to a computing server via cables. After the kernels have been placed, the photograph process can be started by clicking a button, then the light will be turned on and the camera will capture the image and transmit it to the server for analysis.

\subsection{Multi-Classification of Rice Kernel Flaws}
\label{sec:method:class}
\begin{figure}[htbp]
    \centering
    \begin{tabular}{c}
         \includegraphics[width=0.35\textwidth]{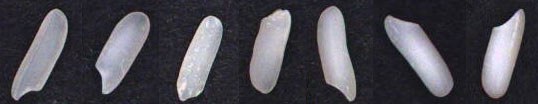}%&
    \end{tabular}
    \caption{Rice kernels with different transparency.}
    \label{fig:chalky}
\end{figure}

\begin{figure}[htbp]
    \centering
    \begin{tabular}{c}
         \includegraphics[width=0.3\textwidth]{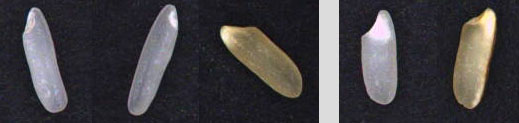}%&
    \end{tabular}
    \caption{Left: normal kernels and yellow kernel with white embryo; Right: chalky kernels.}
    \label{fig:embryo}
\end{figure}

Rice kernels can only be accurately distinguished by professionals in manual inspection work in the past. Traditional image classification methods relying on image features are neither effective for the recognition of rice flaws, as the boundaries between some types of rice kernels are not clear. Fig.~\ref{fig:chalky} shows some rice kernels with different transparency. It is difficult to quantize the value of transparency and classify the rice kernels as chalky based on a given threshold. Different rice varieties have different appearance. Even for the same variety of rice from the same place of origin, the appearances of the kernels, such as the glossiness and transparency, will probably be different if stored in different environments. Fig.~\ref{fig:embryo} shows an example which is difficult to effectively discriminate. According to China Standard GB/T 1354-2018~\cite{gbt1354}, a chalky kernel contains opaque white parts including white belly, white core, and white back. The rice kernels shown on the left side of Fig.~\ref{fig:embryo} also have white opaque portions, but these white opaque portions are embryos, not chalky lumps. Whereas, the rice kernels shown on the right side of Fig.~\ref{fig:embryo} are chalky kernels. This means it is almost impossible to determine whether the rice kernel is chalky based on solely the detected white opaque patches.

To solve these problems, we employ deep neural networks to classify the rice flaws. It is straight forward to use an object detection algorithm, such as Yolo~\cite{yolov5}, to detect rice kernels in the whole image in one step. While the obtained kernel location is accurate, the classification precision may not be high, especially for the partial-chalky and mass-chalky kernels. Inspired by the works~\cite{verma2010image, kaur2013classification} which identify the chalky rice with grayscale images, we designed a multi-stage workflow for rice kernel localization and flaw classification.

\begin{figure*}[htbp]
    \centering
    \begin{tabular}{c}
         \includegraphics[width=0.75\textwidth]{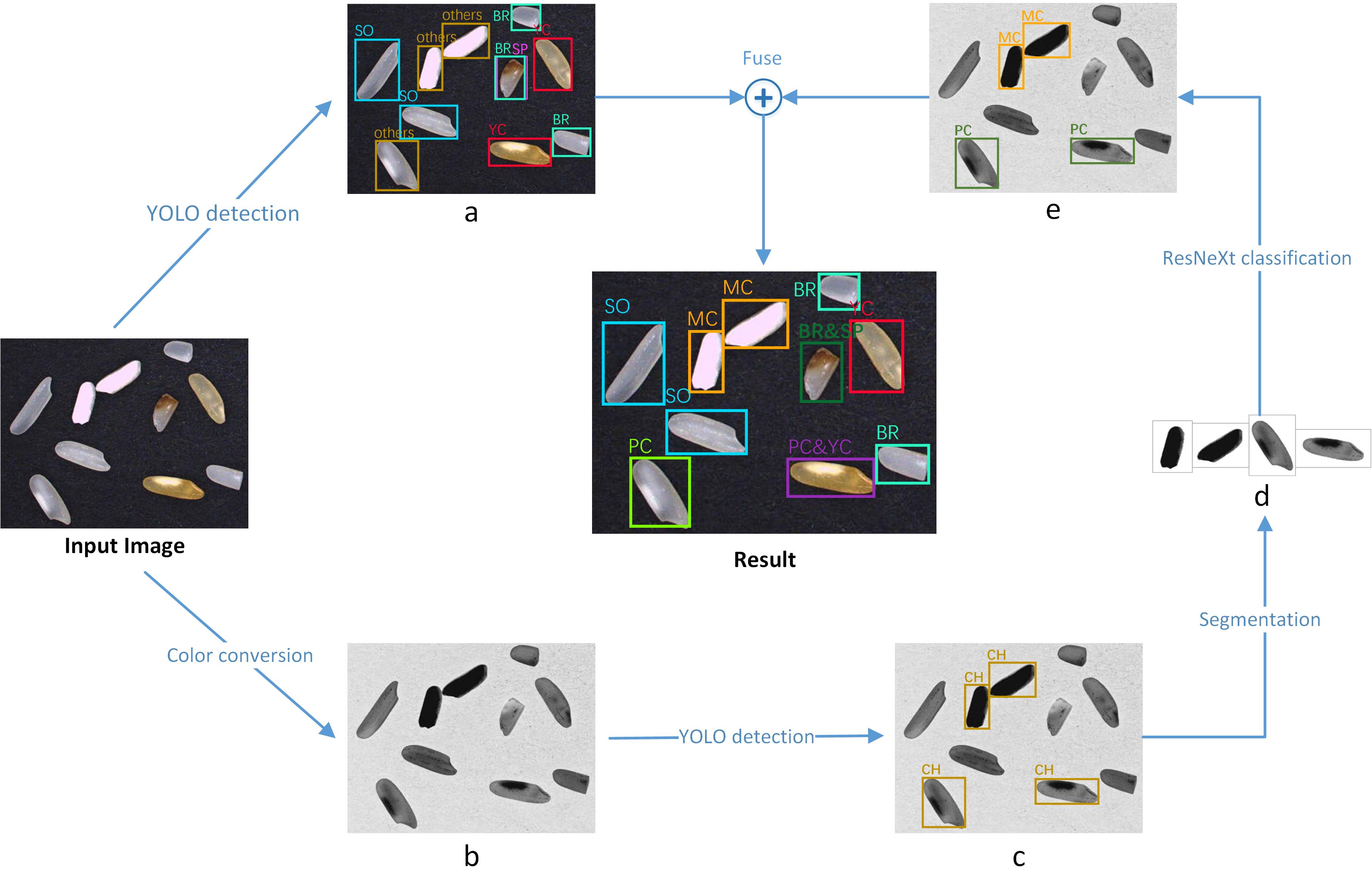}%&
    \end{tabular}
    \caption{Multi-stage workflow for rice kernel detection and rice flaw classification.}
    \label{fig:workflow}
\end{figure*}

This workflow is divided into two branches, as shown in Fig.~\ref{fig:workflow}. One is the detection of YC, SP, BR, and SO kernels. The other is the localization and classification of PC and MC kernels. The output of the two branches are fused together to form the final result. The whole process includes five stages: the detection of YC, SP, BR, and SO kernels, the detection of chalky kernels, segmentation, the classification of PC and MC kernels, and the fusion stage. The details of each stage will be described below.

\textbf{Detection of YC, SP, BR, and SO kernels:} The detection of YC, SP, BR, and SO kernels is implemented by Yolov5~\cite{yolov5}, which shows advantages in both detection accuracy and speed over other methods.

\begin{figure}[htbp]
    \centering
    \begin{tabular}{c}
         \includegraphics[width=0.3\textwidth]{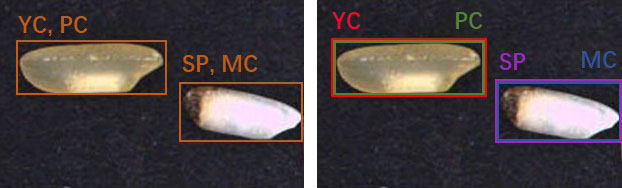}%&
    \end{tabular}
    \caption{Left: each kernel has one bounding box and two classes; Right: each kernel has two bounding boxes, and each bounding box belongs to one category.}
    \label{fig:label}
\end{figure}

As discussed in Section~\ref{sec:method:ps}, the output class of a rice kernel may either be one of the 6 classes with single properties, or those classes with dual properties, and all other possibilities such as a class with a property combination which are not mentioned before should be excluded. This is inherently a multi-classification problem. A Dual-property rice can be labeled as a new category which is independent of the single properties, or as one entity with two categories, or as two entities with different single class. As the number of the samples of dual-property rice is limited, the data imbalance will result in low detection accuracy of dual-property kernels if they are treated as independent categories. Both the other two options seem more suitable comparatively. Since Yolov5 supports multi-classification, annotating a kernel to two categories does not affect the calculation of the loss function, and it can be propagated backward during the network training. On the other hand, Yolov5 predicts multiple bounding boxes per grid cell, so it can detect the same kernel repeatedly. Fig.~\ref{fig:label} illustrates the differences between the annotations of the latter two schemes. On the left side of the figure, both kernels of dual-property have only one bounding box, but each bounding box is annotated as two categories. On the right side, the same two kernels of dual-property are both surrounded by two bounding boxes, each of which corresponds to one class. Here we chose to label dual-property rice kernels as two entities separately considering the convenience of annotation.

The trained Yolo model (denoted as Yolo-1) can locate and classify YC, SP, BR, SO, and other kernels, which is shown in Fig.~\ref{fig:workflow}-a. Rice kernels other than YC, SP, BR, and SO are classified as 'others'. %This hybrid class helps the detection of chalky rice on the other branch.

\begin{figure}[htbp]
    \centering
    \begin{tabular}{c}
         \includegraphics[width=0.3\textwidth]{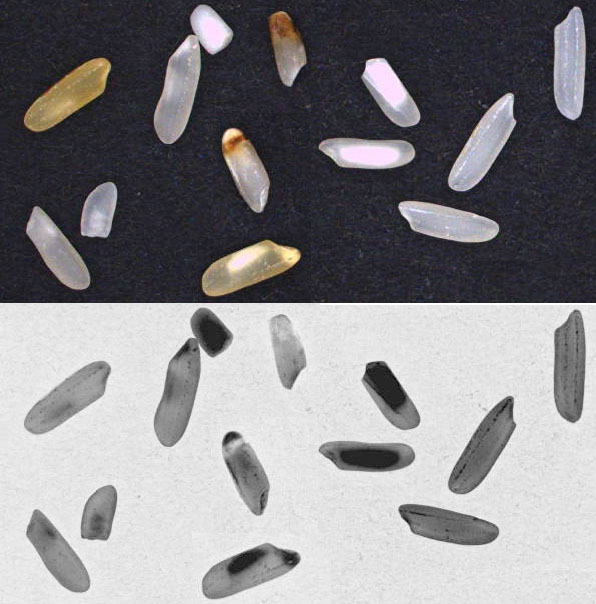}%&
    \end{tabular}
    \caption{Comparison of the original RGB image and our converted gray image.}
    \label{fig:grey}
\end{figure}

\textbf{Detection of chalky kernels:} Fig.~\ref{fig:grey} shows the comparison of a color image and a gray image of rice kernels. There are partial-chalky kernels, mass-chalky kernels, and also chalky kernels with dual properties, such as PC\&YC and PC\&SP kernels. It can be seen that the chalky parts are more obvious in the gray image which completely retains the chalky features and eliminate the occlusion of other features. The gray image in Fig.~\ref{fig:grey} is obtained using the following formula:
\begin{equation}\label{eq:gray}
Y=255-0.299\cdot R+0.587\cdot G+0.114\cdot B,
\end{equation}
where $Y$ is the grayscale value. $R$, $G$, and $B$ are the three channel values of the color image respectively.

Taking the converted gray image as the input of Yolo detection network (Fig.~\ref{fig:workflow}-b), we trained another Yolo model (denoted as Yolo-2), which is responsible for detection chalky kernels in gray images (Fig.~\ref{fig:workflow}-c).

\textbf{Segmentation:} After detecting the chalky kernels from the gray images, we need to segment it out from each detected bounding areas. Here we cutout the pixels of chalky kernels in gray images based on the kernels contours detected in color images, since the boundaries of rice kernels are easier to accurately extract in color images.

%Here we adopted two tricks. One is to locate the chalky kernels using the detection results of the two Yolo models. The 'others' kernels detected by Yolo-1 can be used to compensate for the missed detection of Yolo-2, although this occurs very rarely. The other trick is to cutout the pixels of chalky kernels in gray images based on the kernels contours detected in color images, since the boundaries of rice kernels are easier to accurately extract in color images.

\begin{figure}[htbp]
    \centering
    \begin{tabular}{c}
         \includegraphics[width=0.45\textwidth]{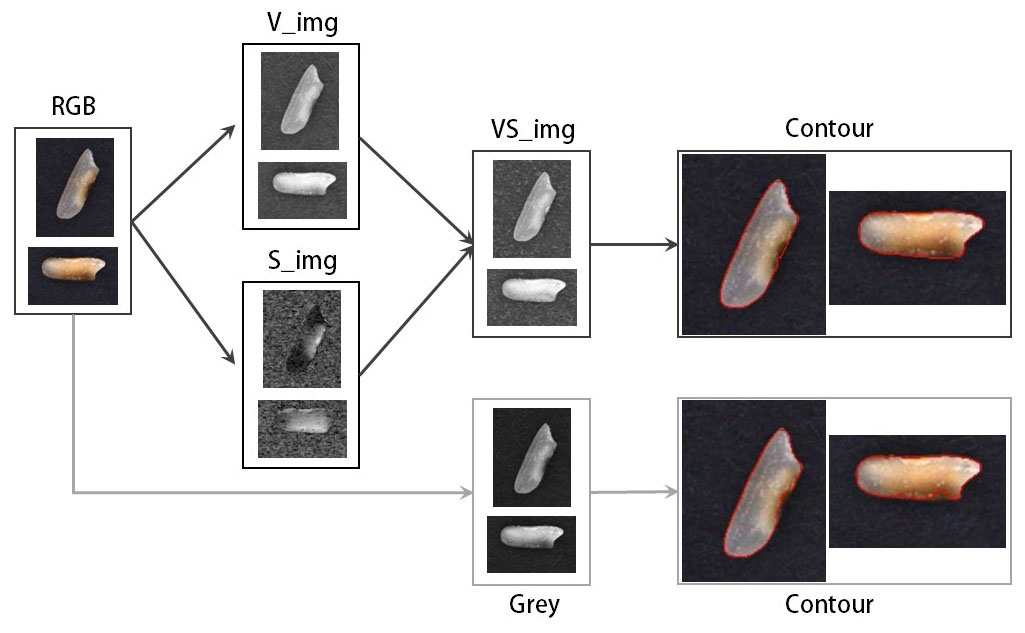}%&
    \end{tabular}
    \caption{Comparison of contours extracted from grey image and our VS\_img.}
    \label{fig:contour}
\end{figure}

As has been introduced, we choose a black material as the photo background as it makes easier to distinguish the kernels. However, this also resulted in the failure of simple contour extraction algorithms to extract the contours of spotted kernels which contain black disease spots. Therefore, we adopt the following processing for kernel contour extraction:
\begin{equation}\label{eq:contour}
\begin{aligned}
&VS\_img=max(erode(S\_img),V\_img),\\
&Contours=findContours(Morph\_Open(VS\_img)),
\end{aligned}
\end{equation}
where $S\_img$ and $V\_img$ are the saturation and hue channels of the image which is converted from RGB color space to HSV color space, and $VS\_img$ is the maximum image of eroded $S\_img$ and $V\_img$. Contours can then be found by applying dilation, erosion and contour extraction operations provided in OpenCV~\cite{opencv}. Fig.~\ref{fig:contour} shows the comparison of the contours of two spotted kernels extracted from gray image and $VS\_img$ respectively. As can be seen, the lesion parts which are missing in the contours extracted with traditional method are contained in the results of our method.

The segmentation is operated as follows: If the center point of the extracted kernel contour is close to a center point of the detected bounding boxes of chalky kernels, we label the pixels inside the contour belongs to a chalky kernel and cutout them to a blank template to construct individual image for each chalky kernel. An example is shown in Fig.~\ref{fig:workflow}-d.

\textbf{Classification of PC and MC kernels:} We employ a classification network, ResNeXt~\cite{Xie2016}, to classify partial-chalky kernels and mass-chalky kernels. It does not require a long time training and the classification accuracy is satisfactory. Besides, its inference is fast. An example output of this stage is shown in Fig.~\ref{fig:workflow}-e.

\textbf{Fusion:} After the above steps, we have detected YC, SP, BR, and SO kernels from color images, and PC and MC kernels from gray images. As these detections are independent of each other, they need to be fused and undefined property combinations should be filtered out. As discussed in Section~\ref{sec:method:ps}, the output class of a rice kernel may either be one of the 6 classes with single properties, or those classes with dual properties, and all other possibilities such as a class with a property combination which are not mentioned before can be excluded.

The fusion operation is processed as follows: For each bounding box, we calculate its IOU (Intersection over Union) with other bounding boxes to get proposals of overlapping bounding boxes. If the bounding box has no overlaps with others, the rice kernel surrounded is classified into the class corresponding to this bounding box, i.e., one of the 6 classes with single properties. If the bounding box overlaps with another bounding box and the combination of the properties corresponding to the two bounding boxes is within the pre-defined range, the rice kernel surrounded by this bounding box is classified into a class with these dual properties. If the bounding box overlaps with more than one bounding box or the combination of the properties corresponding to the overlapped bounding boxes is not defined, we will rank the proposals with their confidence, filter out mutually exclusive terms with low confidence, and obtain the desired properties combination. An example of the fusion result is shown in the center of Fig.~\ref{fig:workflow}.

\subsection{Weight Ratio Estimation}
\label{sec:method:wei}

After classifying each kernel, we need to estimate its weight and compute the ratio of weight for each kernel type. Traditionally, the inspectors weigh the kernels with a high-precision digital steelyard, which is quite tedious and time-consuming. Here we try to estimate the weight ratio using a vision-based approach. A formal way to do that is to get the volume of each kernel and multiply it with the averaged density of such kernel type. However, as the kernel size is quite small and most of its body is transparent, it is difficult to reconstruct it and compute its volume with vision-based methods. We noticed that the shape of the rice kernels is roughly the same except for the broken ones, so we imagined that the rice weight could be estimated in the 2D projection space. Moreover, what matters on the rice quality evaluation is the ratio of the kernel weight. That is, if the weight estimation for different types of rice have similar bias, the weight ratio will not be significantly affected.

Inspired by these observations, we propose to compute the weight ratio of rice kernels via their 2D projected area instead of volume. Specifically, we propose a new weight-per-pixel metric $\rho_t$ ($t\in$ $\{$PC, MC, YC, SP, BR, SO$\}$), which represents the averaged density of rice kernels of type $t$ per pixel, and the relative weight of rice kernels of type $t$ can be obtained by multiplying it with the projected area on the 2D image in pixels. After collecting the weight of all types of rice kernels in this way, the ratios are computed via equation~\ref{eq:ratio}.

To get the area of a kernel, we use the contour detection method mentioned previously and calculate the area inside the kernel contour. It should also be pointed out that as the distance between the lens and the background mat is fixed, the projection matrix is constant, hence the scales of all the kernels are uniform.

To get the weight-per-pixel $\rho_t$, we first collected a large quantity of samples ${K_t}$ for each type of kernels. These samples only consisted of rice kernels with single property. The total weight values $W_t$ of kernels ${K_t}$ for type $t$ are obtained by using a digital steelyard. Then we sum up the areas $A_t$ of ${K_t}$, and the density $\rho_t$ can thus be calculated as:

\begin{equation}\label{eq:rho}
\begin{aligned}
	\rho_t = \frac{W_t}{A_t}.
%	WR_i = &\frac{W_i}{\sum W_i},
%\\& i \in \{PC, MC, YC, US, BR, WH\}.
\end{aligned}
\end{equation}

It is observed that for a certain category of rice, the density fluctuates little. Whereas the density needs to be re-calculated when estimating other rice categories. For dual-property kernels, their weights will be calculated twice, and the results will probably vary as the densities used in the two calculations are different. We take half of each as the weight of the dual-property rice kernel when calculating the total weight of the rice samples. Because dual-property kernels only account for a small portion of all the kernels, this weight error has little effect on the overall result and can therefore be ignored. %For example, the density difference between PC and YC kernels is approximately $3.97e-7 g/pixel$ and the average area of a whole kernel is $3356 pixels$ by our calculations. This means that for a PC&YC kernel, the weight difference between the two estimates is about &1.33e-3 g&.

\section{Experimental Evaluation}
\label{sec:Results}

\subsection{Training Setup}
\label{subsec:training setup}
\textbf{Training data:} Using the hardware setup introduced earlier, we collected a total of 322 images as training data, validation data and test data. Each image contains a variable number of rice kernels, mostly between 40 and 80. All data are annotated by professionals with many years of experience. They are used for the training and testing of our rice detection and classification model.

In order to calculate the density $\rho_t$ for each type of kernels via equation~\ref{eq:rho}, we first invited experts to manually classify some rice samples into different types and weighted them separately with a digital steelyard. We then took photos of these rice samples and calculated their projection area. After that, the density of different types of rice kernels can be calculated. Table~\ref{tb:density} shows the number of kernels we collected, their weighed weight, projection area and the rice density calculated.

\begin{table}[h]
%\footnotesize
\centering
\caption{Rice density calculation.}
%\vspace{-0.5em}
\label{tb:density}
\scalebox{0.9}
{
\begin{tabular}{ccccccc}
\hline
&  \textbf{SO}  & \textbf{PC} &  \textbf{MC} &  \textbf{YC} &  \textbf{SP} &  \textbf{BR}\\
\hline
\begin{tabular}{@{}c@{}}Kernel \\ amount\end{tabular} & 1227      & 952       & 1088       & 810       & 1014      & 1425     \\
\hline
\begin{tabular}{@{}c@{}}Weight \\ (g)\end{tabular} & 21.94   & 17.48   & 15.85    & 9.26    & 10.39   & 12.67  \\
\hline
\begin{tabular}{@{}c@{}}Area \\ (pixels)\end{tabular} & 4.11E6 & 3.22E6 & 2.98E6  & 1.84E6 & 2.1E6 & 2.42E6   \\
\hline
\begin{tabular}{@{}c@{}}Density \\ (g/pixel)\end{tabular}   & 5.32E-6 & 5.42E-6 & 5.31E-6  & 5.03E-6 & 4.93E-6 & 5.23E-6 \\
\hline
\end{tabular}}
\end{table}
%Weight(g) & 21.9459   & 17.4861   & 15.8578    & 9.2693    & 10.3913   & 12.6766  \\
%Area(pixels)       & 4118000.5 & 3222047.5 & 2986299.0  & 1842916.0 & 2105347.5 & 2421457   \\
%Density(g/pixel)   & 5.329E-06 & 5.427E-06 & 5.310E-06  & 5.030E-06 & 4.936E-06 & 5.235E-06 \\

\textbf{Training details:} Our detection and classification networks are implemented with Pytorch~\cite{paszke2019pytorch}. Both the detection model Yolo-1 and Yolo-2 are trained for 300 epochs from the pretrained yolov5x model using SGD optimizer~\cite{bottou2007tradeoffs}, with learning rate of 0.01 and batch size of 16. Then the trained models are fine-tuned for 300 epochs with a small learning rate 0.0032. The size of the input image is 640*640. The classification network ResNeXt is trained for 100 epochs from a pretrained model using Adam optimizer~\cite{kingma2014adam}, with learning rate of 0.001 and batch size of 32. The size of the input image is 224*224.

\subsection{Evaluation}

\textbf{Classification results:} A 5-fold cross validation is applied to test our multi-stage workflow for the multi-classification of rice kernels. The total 322 rice images were shuffled randomly and split into training data (292 images), validation data (15 images), and test data (15 images) for 5 times. We trained and tested our model 5 times with these datasets. Table~\ref{tb:precision} shows the test result, where the ground truth is obtained by letting professional inspectors to classify the kernels one by one. We compare our model with the native Yolo~\cite{yolov5} and Yolo plus postprocessing. This postprocessing refers to the operation of filtering out impossible property combinations from the Yolo detection result, as described in Section~\ref{sec:method:class}.

\begin{table}[h]
%\footnotesize
\centering
\caption{The 5-fold classification results of Yolo, Yolo plus postprocessing and our model.}
%\vspace{-0.5em}
\label{tb:precision}
%\scalebox{0.9}{
{
\begin{tabular}{ccccccc}
\hline
\textbf{Precision($\%$)}& \textbf{SO} & \textbf{BR} & \textbf{PC} &  \textbf{MC} &  \textbf{YC} &  \textbf{SP} \\
\hline
Yolo & 88.20 & 93.77 & 75.76 & 91.11 & 93.22 & 88.35 \\
%\hline
Yolo+p & 92.28 & 96.34 & 81.63 & \textbf{95.89} & 94.86 & 90.49 \\
%\hline
Our model & \textbf{94.25} & \textbf{97.83} & \textbf{86.58} & 86.99 & \textbf{97.61} & \textbf{93.07} \\
\hline
\textbf{Recall($\%$)}& \textbf{SO} & \textbf{BR} & \textbf{PC} &  \textbf{MC} &  \textbf{YC} &  \textbf{SP} \\
\hline
Yolo & \textbf{97.43} & \textbf{96.65} & \textbf{85.75} & 82.22 & 94.87 & \textbf{86.72} \\
%\hline
Yolo+p & 95.27 & 95.88 & 81.27 & 78.07 & 94.67 & 85.98 \\
%\hline
Our model & 95.35 & 95.21 & 83.38 & \textbf{96.52} & \textbf{96.84} & \textbf{86.72} \\
\hline
\textbf{F1 Score}& \textbf{SO} & \textbf{BR} & \textbf{PC} &  \textbf{MC} &  \textbf{YC} &  \textbf{SP} \\
\hline
Yolo & 92.58 & 95.19 & 80.45 & 86.44 & 94.04 & 87.52 \\
%\hline
Yolo+p & 93.75 & 96.11 & 81.45 & 86.07 & 94.77 & 88.17 \\
%\hline
Our model & \textbf{94.80} & \textbf{96.50} & \textbf{84.95} & \textbf{91.51} & \textbf{97.23} & \textbf{89.78} \\
\hline
\end{tabular}}
\end{table}

We calculate the classification precision, recall and F1 score of the six types of rice kernels. The dual-property kernels are counted into each single type. As can be seen from Table~\ref{tb:precision}, the average precision of our model for the six types are all larger than 85$\%$. The precision for sound kernels, broken kernels, yellow-colored kernels, and spotted kernels are over 90$\%$. Compared with Yolo and Yolo plus postprocessing, our model performs the best in all types of rice classification. Although the classification precision of mass-chalky kernel is lower than other models, the recall rate is improved, and the F1 score is still the highest.

\textbf{Weight estimation results:}
Five sets of rice samples are used to test the effectiveness of the weight estimation method proposed in this paper. Each set contains a different amount of kernels in different types. First, the rice samples were classified using our trained multi-stage model. Then we estimated the weight of each type of rice kernels with their projected area and density. At last, the weight ratios were calculated according to equation~\ref{eq:ratio}. The rice samples have been sorted and precisely weighed by professionals, and we will compare our calculation result with these expert data.

\begin{table}[h]
%\footnotesize
\centering
\caption{Test results of weight ratio of defective rice.}
%\vspace{-0.5em}
\label{tb:weight}
%\scalebox{0.9}{
{
\begin{tabular}{cccccc}
\hline
& \textbf{BR}  & \textbf{PC} &  \textbf{MC} &  \textbf{YC} &  \textbf{SP} \\
\hline
mean error($\%$) & 1.8926 & 1.4620 & 1.9038 & 0.4108 & 1.7482 \\
max error($\%$) & 2.4514 & 2.7315 & 2.9977 & 1.4765 & 2.4498 \\
\hline
\end{tabular}}
\end{table}

Table \ref{tb:weight} shows the evaluation result. It can be seen that all mean errors are less than 2$\%$, and the maximum errors are not greater than 3$\%$. In comparison, the error of the mass-chalky kernel is the largest, and the error of yellow-colored kernel is the smallest.

\begin{figure*}[t]
    \centering
    \begin{tabular}{c@{\hskip3pt}c@{\hskip3pt}c}
         \includegraphics[width=0.95\textwidth]{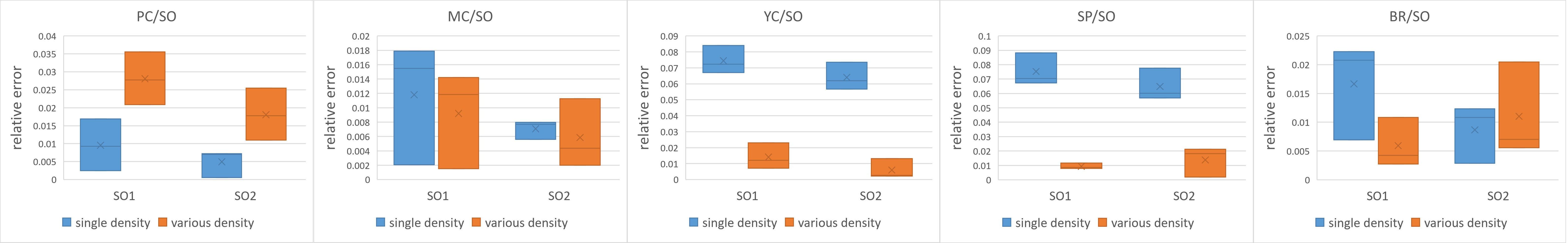}
    \end{tabular}
    \caption{Box diagrams of the relative error of estimated weight percentage calculated with a single density and various densities. PC, MC, YC, SP, BR and SO represent Partial-Chalky, Mass-Chalky, Yellow-Colored, Spotted, Broken and Sound kernels respectively.}
    \label{fig:density}
\end{figure*}

\subsection{Discussion}
\textbf{Single density vs. various density: } According to our intuition, broken kernels and sound kernels have the same physical properties but different sizes, so their densities should be the same. But on the 2D projection surface, broken kernel of the same weight should have more edge parts than whole kernel, so the projected density of broken kernels should be less than that of the whole kernels. Chalky, yellow-colored and spotted kernels all have character changes, so their densities should be different from sound rice, and their 2D projection densities should also be different. But how much do the projected densities of different types of rice kernels differ? Can the same density be used to estimate the weight ratios for these types of rice kernels? To answer these questions, we conducted a test.

Some rice samples classified by experts are used for this test. We randomly divided each of the five types of defective rices into three groups. The remaining sound rice was divided into two groups. They occupied the majority of all rice samples. Then the accurate weight values, projected area values and estimated weight were obtained respectively. Table~\ref{tb:densitytest} shows the detail.

\begin{table}[h]
%\footnotesize
\centering
\caption{The rice samples used for density test.}
%\vspace{-0.5em}
\label{tb:densitytest}
{
\begin{tabular}{ccccc}
\hline
& \begin{tabular}{@{}c@{}}\textbf{kernel}\\\textbf{amount}\end{tabular} & \begin{tabular}{@{}c@{}}\textbf{accurate}\\\textbf{weight}(g)\end{tabular}
& \begin{tabular}{@{}c@{}}\textbf{area}\\(pixels)\end{tabular} & \begin{tabular}{@{}c@{}}\textbf{estimated}\\\textbf{weight}(g)\end{tabular}\\
\hline
SO 1 & 148 & 2.6927 & 494365 & 2.6346 \\
SO 2 & 338 & 5.9928 & 1111058& 5.9211 \\
\hline
PC 1 & 97 & 1.7933 & 330068.5 & 1.7913 \\
PC 2 & 109 & 1.9313 & 360577.5 & 1.9569 \\
PC 3 & 127 & 2.3933 & 443462.5 & 2.4067 \\
\hline
MC 1 & 111 & 1.8700 & 348635.5 & 1.8513 \\
MC 2 & 115 & 1.9726 & 362904.5 & 1.9271 \\
MC 3 & 133 & 2.2357 & 417796.5 & 2.2186 \\
\hline
YC 1 & 134 & 1.6182 & 322061.5 & 1.6199 \\
YC 2 & 149 & 1.8479 & 363798.5 & 1.8298 \\
YC 3 & 144 & 1.8580 & 364007 & 1.8308 \\
\hline
SP 1 & 109 & 1.3254 & 260486.5 & 1.2857 \\
SP 2 & 139 & 1.6427 & 328200.5 & 1.6199 \\
SP 3 & 144 & 1.7985 & 352404.5 & 1.7394 \\
\hline
BR 1 & 98  & 0.9846 & 182024 & 0.9529 \\
BR 2 & 129 & 1.2161 & 227903 & 1.1931 \\
BR 3 & 137 & 1.3338 & 250334 & 1.3105 \\
\hline
\end{tabular}}
\end{table}

If a single density value is used, the weight ratio is equal to the area ratio. We calculate the accurate weight percentage, area percentage and estimated weight percentage of the five types of defective kernels relative to the sound kernels, where the area percentage is equal to the estimated weight percentage when different types of rice kernels are supposed to have the same density. In order to analyze the effect of rice density on rice quality estimation, we compare the relative error of area percentage relative to accurate weight percentage and the relative error of estimated weight percentage relative to accurate weight percentage.

Figure ~\ref{fig:density} shows the box diagrams of the relative error calculated for the five types of defective rice. As can be seen from the chart, the error median line using the same density is higher than the error median line using respective densities for mass-chalky kernels, yellow-colored kernels, spotted kernels and broken kernels. Only the partial-chalky kernels have different result from our expectation, which means they are more suitable to apply the same density as sound kernels. We infer the reason is that the density of PC is very close to that of SO, since the chalky part in PC is small. Therefore, the error of the area ratio is smaller, and using an inaccurate projection density will introduce greater deviation. %The specific data of the test result can be found in the appendix.

\section{Conclusion}
\label{sec:Con}

In this paper, we propose an integrated system for rice quality estimation. A multi-stage workflow is employed to locate and classify the rice kernels with possibly overlapped types of flaws. A density per pixel metric is proposed to measure the weight ratios of kernels with different types, such that the quality estimation can be performed with a fully vision-based approach. We performed various experiments to show the advantage and usefulness of our system and proved that rice quality estimation can be carried out automatically and tedious manual works can be replaced.

%\section{Reference Examples}

%\begin{thebibliography}{00}

%\bibitem{b1} G. O. Young, ``Synthetic structure of industrial plastics,'' in \emph{Plastics,} 2\textsuperscript{nd} ed., vol. 3, J. Peters, Ed. New York, NY, USA: McGraw-Hill, 1964, pp. 15--64.

%\end{thebibliography}
\bibliographystyle{IEEEtran}
\bibliography{references}

\end{document}